\documentclass[letterpaper, 10 pt, conference]{ieeeconf}  

\IEEEoverridecommandlockouts                              

\usepackage{graphics} 
\usepackage{epsfig} 
\usepackage{times} 
\usepackage{amsmath} 
\usepackage{amssymb}  

\usepackage{booktabs}
\usepackage{multirow}

\newtheorem{assumption}{Assumption}
\newtheorem{theorem}{Theorem}

\title{\LARGE \bf
Preference-Augmented Distributional Reinforcement Learning for Risk-Sensitive Navigation
}

\author{Zhaofan Zhang$^{1}$, Yue Chang$^{1}$, Rufeng Chen$^{1}$, Hechang Chen$^{3}$, Sihong Xie$^{1 \dagger}$, Hui Xiong$^{1,2 \dagger}$
\thanks{$^{1}$Thrust of Artificial Intelligence, The Hong Kong University of Science and Technology (Guangzhou),
China. $^{2}$Department of Computer Science and Engineering, The Hong Kong University of Science and Technology Hong Kong SAR, China. $^{3}$School of Artificial Intelligence, Jilin University, China.} 
\thanks{$\dagger$Corresponding authors.}
}

\begin{document}

\maketitle
\thispagestyle{empty}
\pagestyle{empty}

\begin{abstract}

Ensuring safe and reliable navigation under diverse sources of uncertainty remains a central challenge for autonomous systems.
Within distributional reinforcement learning (DistRL), Implicit Quantile
Networks (IQN) have attracted interest for navigation under
uncertainty because they enable risk adaptation through flexible
quantile sampling. Existing approaches, however, typically use a fixed or
randomly sampled Conditional Value-at-Risk (CVaR) level during training and
adjust the risk level during deployment using hand-tuned hyperparameters or
external sensor signals.
This scalar-CVaR formulation confines learning to a narrow family of tail-risk
preferences, creating a mismatch between training-time preference exposure and
deployment-time risk adjustment. We instead represent variations in risk
preference as shifts in quantile-weight distributions and expose the agent to
diverse preferences through preference-guided exploration. At deployment, we
optimize against the worst-case preference distribution to derive a
conservative policy that accounts for multiple risk profiles. This reduces
hyperparameter tuning and reliance on limited sensor signals under partial
observability.
Experiments across simulated environments and vehicle platforms show that our
method reduces collisions while improving task success.

\end{abstract}


\section{Introduction}

Autonomous navigation in unknown environments requires agents to make risk-sensitive decisions from partial observations while coping with latent environmental uncertainty, such as airflow, water currents, and occlusions~\cite{nahavandi2025comprehensive,lee2023adaptive,wang2024research}. Such uncertainty is often only partially observable, such that similar observations may correspond to markedly different latent dynamics and future outcomes, making it difficult to determine an appropriate degree of policy conservatism from the current observation alone.

To account for the variability in future outcomes induced by such uncertainty, distributional reinforcement learning (DistRL)~\cite{bellemare2017distributional,dabney2018distributional} models the distribution of returns rather than only their expectation. This distributional representation enables decisions to account for adverse outcomes beyond the expected return and has motivated risk-sensitive applications in robotic tasks such as pursuit-evasion~\cite{zhang2025terl}, quadruped locomotion~\cite{schneider2024learning,shi2024robust}, and autonomous navigation~\cite{liu2022adaptive,lin2023robust,zhang2025perturbation,zhang2026quantile}. A common approach is to apply Conditional Value-at-Risk (CVaR) to the learned return distribution at inference time. For return maximization, a smaller CVaR level restricts action evaluation to lower-tail quantiles, placing greater emphasis on adverse outcomes and inducing more conservative behavior. We refer more generally to how different parts of the return distribution are weighted during action evaluation as the agent's \emph{risk preference}.
Among DistRL methods, Implicit Quantile Networks (IQN)~\cite{dabney2018implicit} provide a flexible representation of the return distribution by conditioning value estimates on continuously sampled quantile fractions. 
This enables risk preferences to be expressed through distortions over the quantile domain, with CVaR a widely used choice for risk-sensitive action selection.

Existing robotic applications typically impose different risk preferences during training and deployment. During training, prior work either adopts risk-neutral action selection over the full return distribution~\cite{lin2023robust,lin2024decentralized} or introduces preference diversity by sampling a CVaR level at the beginning of each episode and keeping it fixed throughout the trajectory~\cite{kamran2021minimizing,liu2022adaptive}. At deployment, risk-sensitive action selection is commonly performed using a fixed CVaR level selected for the desired degree of conservativeness~\cite{kamran2021minimizing}, or by dynamically adjusting the CVaR level according to observable risk indicators, such as obstacle proximity~\cite{lin2023robust} or perception-derived risk estimates~\cite{liu2022adaptive,zhang2025path,sun2026dynamic}.

\begin{figure}[t]
\centering  
\vspace{0.2cm}
\includegraphics[width=0.98\linewidth]{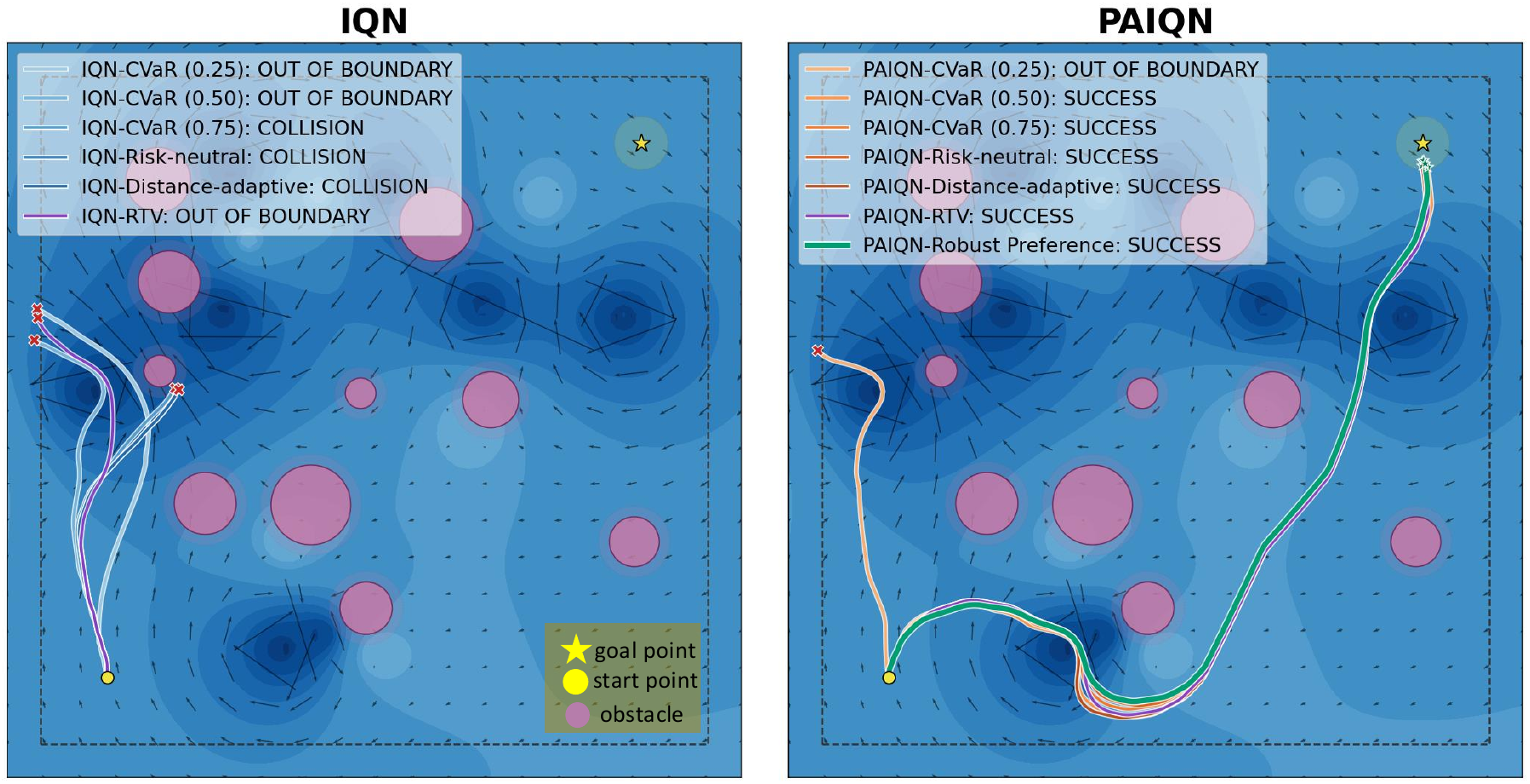}
\caption{\textbf{Proposed PAIQN \textsc{vs.} IQN:} ASV trajectories under different risk preferences in a marine environment with obstacles and shadowed vortices. PAIQN achieves successful navigation under most evaluated preference settings, whereas IQN consistently encounters collisions or boundary violations, indicating improved preference robustness and success across diverse risk preferences.}
\vspace{-0.4cm}
\label{fig1}
\end{figure}

This separation creates a mismatch between training-time preference exposure and deployment-time risk adaptation. As the agent moves through an environment, the risk associated with different states can change substantially, requiring its degree of conservativeness to vary accordingly. Yet existing training schemes expose the policy either to a single risk-neutral preference or to a risk preference that remains fixed within each trajectory. The policy therefore receives limited experience in making sequential decisions under changing risk preferences, despite such adaptation being required during deployment.
The use of CVaR further constrains this adaptation to a specific family of tail-focused preferences. For a given CVaR level, action evaluation depends only on the selected lower-tail quantiles, while the remaining quantiles contribute nothing to the decision. This discards information represented by the full return distribution and limits the range of risk preferences available to the policy. Moreover, using a fixed CVaR at deployment offers limited adaptability across changing conditions, whereas dynamically selecting CVaR from external signals relies on predefined mappings and assumes access to sufficiently informative risk indicators, which may be difficult under partial observability. These limitations motivate a more general formulation that exposes the policy to varying risk preferences during training and supports robust adaptation over the full return distribution at deployment.

To address these limitations, we take a broader distributional view of risk preference. Rather than parameterizing preference solely through a scalar distortion parameter, we represent it through the weights assigned across return quantiles. Different quantile-weight profiles emphasize different regions of the learned return distribution, enabling a richer family of preferences than the tail truncations induced by varying a single CVaR level. This representation also provides a natural way to connect training and deployment: during training, varying quantile weights can guide action selection under diverse preferences, while at deployment, the appropriate preference can be treated as uncertain rather than inferred from a fixed CVaR level or a predefined sensory rule.

Building on this perspective, we propose \emph{Preference-Augmented Implicit Quantile Networks} (PAIQN) for risk-sensitive navigation as illustrated in Fig.~\ref{fig1}. During training, PAIQN samples quantile-weight distributions and employs progressively annealed preference-guided exploration, incorporating preference variation into individual action decisions while gradually reducing exploration as learning proceeds. At deployment, we formulate risk adaptation as a distributionally robust optimization problem over an ambiguity set of quantile-weight distributions and evaluate actions against the worst-case preference. This enables conservative decision-making across plausible risk profiles without requiring a predefined CVaR--risk-sensitivity mapping rule, while connecting diverse preference exposure during training with robust preference adaptation at deployment.

Our contributions are summarized as follows:
\begin{itemize}
\item We propose a preference-aware DistRL method for risk-sensitive navigation that represents risk preferences through quantile-weight distributions and incorporates them into progressively annealed preference-guided exploration during training, exposing the agent to more diverse risk-sensitive decisions than varying a scalar CVaR level.
\item We formulate deployment-time risk adaptation as a distributionally robust optimization problem over an ambiguity set of quantile-weight distributions, enabling conservative action evaluation against adverse preference variations.
\item We evaluate the proposed method across multiple simulated navigation environments and vehicle platforms, demonstrating reduced collision rates and improved task success under diverse environmental uncertainties.

\end{itemize}

\section{Related Work}

\subsection{Training-Time Risk Preferences in DistRL Policies}

Distributional reinforcement learning provides a natural basis for
risk-sensitive decision-making by allowing different regions of the return
distribution to be emphasized during action evaluation.
Early robotic applications exploit this property by applying predefined
distortion measures to the learned return distribution.
For example, Quantile QT-Opt~\cite{bodnar2019quantile} investigates different
risk distortions for risk-aware robotic control, while
Kamran et al.~\cite{kamran2021minimizing} use distributional value estimates
to regulate the conservativeness of learned driving policies.
Beyond using a fixed risk criterion, subsequent approaches expose the policy
to multiple risk preferences during training.
RC-DSAC~\cite{choi2021risk} conditions policy learning and value estimation
on a risk parameter, while Distributional PPO~\cite{schneider2024learning}
samples varying risk parameters to obtain behaviors ranging from risk-averse
to risk-seeking.
In navigation, ART-IQN~\cite{liu2022adaptive} further introduces varying
CVaR levels during training, whereas Lin et al.~\cite{lin2023robust} retain
risk-neutral IQN training before adapting risk sensitivity at deployment.
Despite these advances, existing training schemes typically maintain one
risk preference over an episode or restrict preference variation to a scalar
parameter within a predefined distortion family. This provides limited
exposure to preference transitions within a trajectory and restricts the
diversity of distributional preferences encountered during learning.
In this work, we instead represent risk preferences through quantile-weight
distributions and progressively vary them during training to guide
within-trajectory exploration under diverse preferences.

\subsection{Deployment-Time Risk Preference Adaptation}

While training-time methods determine how risk preferences are incorporated
into policy learning, deployment-time methods focus on adapting these
preferences as environmental conditions change.
A straightforward strategy is to use a fixed risk parameter selected for the
desired level of conservativeness, as in risk-conditioned
policies~\cite{choi2021risk,schneider2024learning,shi2024robust}.
More recent navigation methods adapt risk sensitivity online.
ART-IQN~\cite{liu2022adaptive} estimates uncertainty from the lower tail of
the predicted return distribution and adjusts the CVaR level through
exponentially weighted forecasting.
Similarly, Lin et al.~\cite{lin2023robust} adapt CVaR-based action evaluation
according to obstacle proximity, increasing conservativeness as hazards
become more immediate.
These approaches demonstrate that risk sensitivity can be adjusted during
deployment without retraining the underlying distributional value function.
However, fixed risk settings provide limited adaptability, while dynamic
methods depend on predefined mappings from selected uncertainty statistics
or task-dependent sensory cues to a scalar risk level. Under partial
observability, such signals may provide incomplete evidence of latent risk.
In this work, we instead treat quantile weighting itself as uncertain and
perform robust action evaluation over an ambiguity set of preference
distributions, avoiding an explicit mapping from individual risk indicators
to a CVaR level.

\section{Methodology}

\begin{figure}[t]
\centering  
\vspace{0.2cm}
\includegraphics[width=0.98\linewidth]{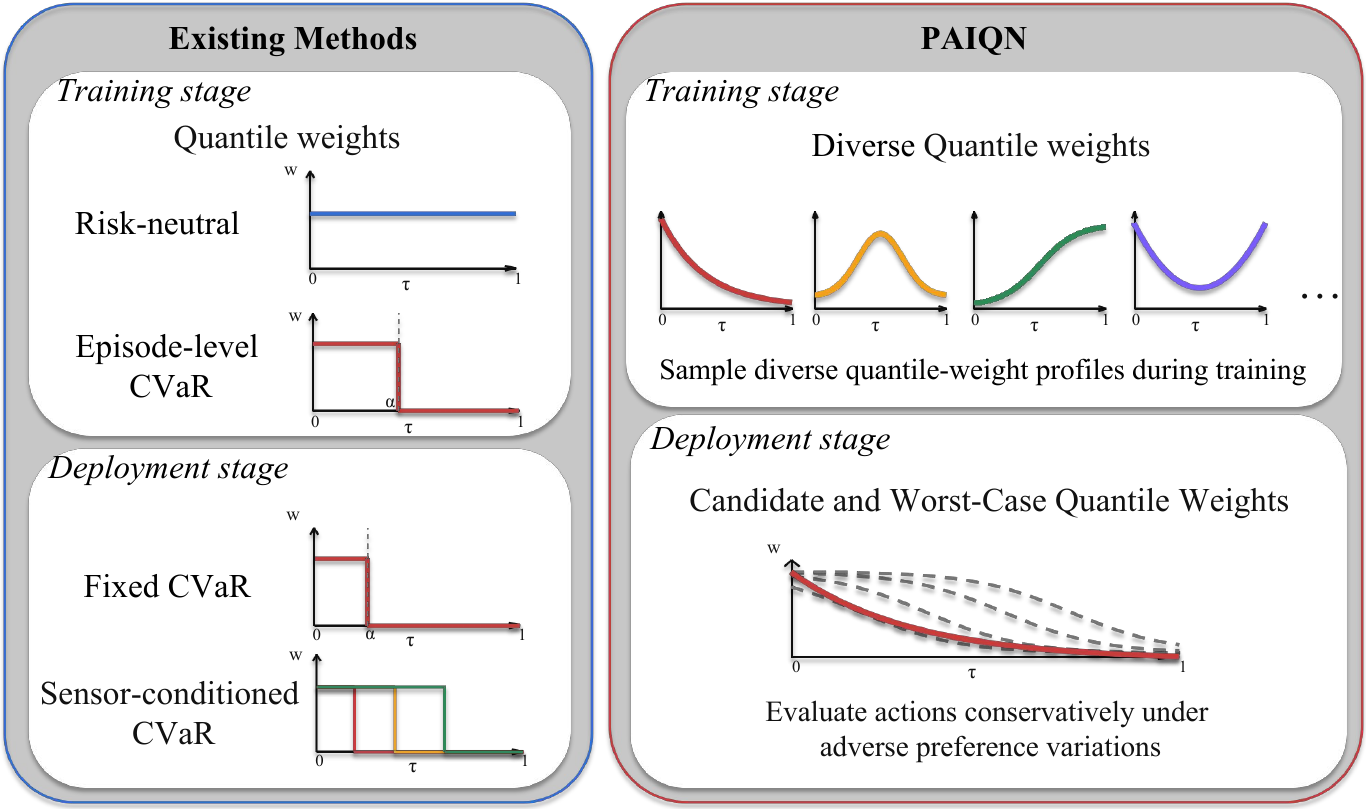}
\caption{PAIQN quantile weights for risk preference.}
\vspace{-0.4cm}
\label{fig2}
\end{figure}

In this work, we develop a preference-aware distributional reinforcement
learning framework for risk-sensitive navigation. We first introduce DistRL
and IQN (Section~\ref{subsec:distrl}), followed by a quantile-weight representation of risk
preferences (Section~\ref{subsec:riskprefer}). We then propose preference-augmented IQN (PAIQN)
for training, exposing the agent to diverse preference-dependent action
evaluations, with representative training-time preference profiles illustrated
in Fig.~\ref{fig2} (Section~\ref{subsec:PAIQN}). Finally, we introduce a
preference-robust mechanism for conservative action evaluation during
deployment (Section~\ref{subsec:deployment}).

\subsection{Distributional Reinforcement Learning}\label{subsec:distrl}

Let \(G_t=\sum_{k=0}^{\infty}\gamma^k r_{t+k}\) denote the discounted return from time step \(t\). Conventional value-based reinforcement learning summarizes this random return by its conditional expectation
\(Q^\pi(s,a)=\mathbb{E}[G_t\mid s_t=s,a_t=a]\).
Consequently, different outcome distributions may become indistinguishable if they share the same expectation. DistRL instead represents the return itself as a random variable \(Z^\pi(s,a)\), whose expectation recovers the conventional action value \(
Q^\pi(s,a)=\mathbb{E}\!\left[Z^\pi(s,a)\right]
\).
For a fixed policy \(\pi\), the return distribution satisfies the distributional Bellman equation
\begin{equation}
\label{zdef}
Z^\pi(s,a)
\stackrel{D}{=}
R(s,a)+\gamma Z^\pi(S',A'),
\end{equation}
where \(S'\sim P(\cdot\mid s,a)\),
\(A'\sim\pi(\cdot\mid S')\), and
\(\stackrel{D}{=}\) denotes equality in distribution.
For value-based control, the corresponding distributional optimality operator is
\begin{equation}
\label{operator}
(\mathcal{T}Z)(s,a)
\stackrel{D}{=}
R(s,a)+\gamma Z\!\left(S',a_Z^\star(S')\right),
\end{equation}
where
\(
a_Z^\star(s')
\in
\operatorname*{argmax}_{b}
\mathbb{E}\!\left[Z(s',b)\right]
\).
Retaining the full return distribution makes properties such as dispersion, asymmetry, and tail outcomes available for decision making.

We approximate \(Z(s,a)\) using Implicit Quantile Networks (IQN)~\cite{dabney2018implicit}.
Rather than assigning probabilities to a predefined support, IQN implicitly approximates the inverse cumulative distribution function
$
Z_\tau(s,a)=F^{-1}_{Z(s,a)}(\tau)
$
by conditioning the value estimator on continuously sampled quantile levels \(\tau\sim\mathcal{U}([0,1])\), thereby allowing return quantiles to be evaluated at arbitrary fractions during both training and deployment.



Given prediction
quantiles $\tau_i\sim\mathcal{U}([0,1])$ and target quantiles
$\tau'_j\sim\mathcal{U}([0,1])$, the pairwise temporal-difference error of IQN is
$\delta_{ij}
=
r+\gamma Z_{\tau'_j}(s',a_\beta^\star(s'))
-Z_{\tau_i}(s,a)$.
The network is optimized using \begin{equation}\label{iqnloss}
\mathcal{L}_{IQN}(s,a,r,s')=\frac{1}{N'}\sum_{i=1}^N\sum_{j=1}^{N'} \rho_{\tau_i}^{\kappa}\!\left(\delta^{\tau_i,\tau'_j}\right),
\end{equation} 
where $\rho_{\tau_i}^{\kappa}$ denotes the quantile Huber loss
\cite{huber1992robust}, which retains the asymmetric quantile-regression
weighting while smoothing small TD errors for more stable optimization.


\subsection{Risk Preferences as Quantile Weights}\label{subsec:riskprefer}
Risk preferences in IQN can be incorporated by applying a distortion function
$\beta:[0,1]\rightarrow[0,1]$ to the sampled quantile levels. The
corresponding risk-sensitive action value is defined as
\begin{equation}
\label{qtoz}
Q_\beta(s,a)
=
\mathbb{E}_{\tau\sim\mathcal{U}([0,1])}
\left[
Z_{\beta(\tau)}(s,a)
\right].
\end{equation}
A widely used risk measure in recent work~\cite{liu2022adaptive,lin2023robust,shi2024robust} is the lower-tail Conditional Value-at-Risk (CVaR)
at level $\alpha\in(0,1]$, induced by the distortion
$\beta_\alpha(\tau)=\alpha\tau$. The resulting action value can be written as
\begin{equation}
Q_{\mathrm{CVaR}_\alpha}(s,a)
=
\frac{1}{\alpha}
\int_{0}^{\alpha}
Z_u(s,a)\,\mathrm{d}u.
\end{equation}
When $\alpha=1$, this formulation reduces to the risk-neutral expected
return, whereas decreasing $\alpha$ places progressively greater emphasis
on unfavorable outcomes in the lower tail of the return distribution.
In practice, given $K$ independently sampled quantile fractions
$\tilde{\tau}_k\sim\mathcal{U}([0,1])$, the corresponding greedy action can
be approximated using the Monte Carlo estimate
\(
\label{policy}
a_\beta^\star(s)
=
\operatorname*{argmax}_{a}
\frac{1}{K}
\sum_{k=1}^{K}
Z_{\beta(\tilde{\tau}_k)}(s,a).
\)

This scalar parameterization is only one way to express preference. For
notational simplicity, we suppress function-approximation parameters and write
$Z_\tau(s,a)$ for the learned return-quantile estimate throughout. Consider
$K$ fixed quantile levels
$\tau_k=(k-\tfrac{1}{2})/K$, $k=1,\ldots,K$. Let $w_k$ be the nonnegative
weight assigned to $Z_{\tau_k}(s,a)$, with
$\sum_{k=1}^{K}w_k=1$. The corresponding action value is the weighted average
\begin{equation}
Q_w(s,a)=\sum_{k=1}^{K}w_k Z_{\tau_k}(s,a).
\label{eq:weighted_quantiles}
\end{equation}
Uniform weights $w_k=1/K$ approximate the risk-neutral expectation, while
discretized CVaR assigns weight only to quantiles in the selected lower tail.
Allowing the individual $w_k$ to vary is more general than changing a single
CVaR level and can emphasize different regions of the predicted return
distribution.



\subsection{Preference-Augmented IQN Training}\label{subsec:PAIQN}


Let $t\geq0$ denote the global interaction step, which is not reset between
episodes. At each preference-based decision, we sample
$\mathbf{x}_t=(x_{t,1},\ldots,x_{t,K})\sim
\operatorname{Dir}(b\mathbf{1}_K)$ and construct the quantile weights as
\begin{equation}
\begin{aligned}
\Delta_t&=\Delta_0\max(t,1)^{-(1+\eta)}, \qquad \eta>0,\\
\widetilde w_{t,k}&=\max\!\left\{1+\Delta_t(Kx_{t,k}-1),0\right\},\\
w_{t,k}&=\frac{\widetilde w_{t,k}}
{\sum_{j=1}^{K}\widetilde w_{t,j}}.
\end{aligned}
\label{eq:training_weights}
\end{equation}
The Dirichlet concentration $b$ controls the
shape of the sampled preference, while $\Delta_t$ controls its deviation
from uniform weighting. The normalization ensures $w_{t,k}\geq0$ and
$\sum_{k=1}^{K}w_{t,k}=1$. The same preference weights are shared across all
candidate actions at a state, yielding the preference-greedy action
\begin{equation}
\hat a_t=
\operatorname*{argmax}_{a\in\mathcal A}
\sum_{k=1}^{K}
w_{t,k}Z_{\tau_k}(s_t,a).
\label{eq:preference_greedy}
\end{equation}
As $\Delta_t\rightarrow0$, Eq.~\eqref{eq:training_weights} approaches
uniform quantile weighting. Hence, preference perturbations are strongest
early in training and gradually vanish, allowing the agent to explore
actions favored by diverse return-distribution preferences without
introducing a persistent directional bias.

Preference-guided exploration is combined with conventional random-action
exploration~\cite{watkins1992q,Mnih2015HumanlevelCT}. With probability $\epsilon_t$, the behavior policy samples an
action uniformly from $\mathcal A$; otherwise, it selects
$\hat a_t$ from Eq.~\eqref{eq:preference_greedy}. We use the linear schedule
\begin{equation}
\epsilon_t=\epsilon_0+
\min\!\left(\frac{t}{f_\epsilon T},1\right)
(\epsilon_f-\epsilon_0),
\label{eq:epsilon_schedule}
\end{equation}
where $T$ is the training horizon and $f_\epsilon$ is the annealing fraction.
The random-action decision is made before preference sampling, so
Eq.~\eqref{eq:training_weights} is evaluated only when the
preference-guided branch is taken.

The same preference mechanism is used when constructing the Bellman target.
For each sampled next state $s'$, we independently draw
$\mathbf{w}'_t$ according to Eq.~\eqref{eq:training_weights} and select
\begin{equation}
\hat a_t^{\mathrm{tar}}(s')
=
\operatorname*{argmax}_{a'\in\mathcal A}
\sum_{k=1}^{K}
w'_{t,k}Z_{\tau_k}(s',a').
\label{eq:preference_target_action}
\end{equation}
After the target action is determined, its return quantiles are used in the
standard IQN Bellman target without preference reweighting. Thus, the
preference mechanism modifies behavior and Bellman target-action selection,
while leaving both the target return distribution and the quantile Huber
objective unchanged.

We next analyze whether the preference-dependent action selection introduces a persistent deviation
from the underlying risk-neutral Bellman optimality update.

\begin{assumption}[Idealized Bellman Setting]
\label{ass:bellman}
We consider a discounted MDP with a Markov state representation, a finite
action set $\mathcal A$, and bounded rewards
$r\in[r_{\min},r_{\max}]$. The return distribution is represented exactly
by its quantile function, and Bellman updates are performed synchronously.
Action evaluation follows the $K$ midpoint quantiles used in
Eqs.~\eqref{eq:preference_greedy} and
\eqref{eq:preference_target_action}, with
$\tau_k=(k-\frac{1}{2})/K$.
\end{assumption}

Under Assumption~\ref{ass:bellman}, let
\(
D=\frac{r_{\max}-r_{\min}}{1-\gamma}\) and
\(Q_t(s,a)=\mathbb{E}[Z_t(s,a)],
\)
where $Q_t(s,a)$ denotes the risk-neutral mean action value induced by the
return distribution at Bellman iteration $t$. Let $Q^\star$ denote the
unique fixed point of the standard scalar Bellman optimality operator.

\begin{theorem}[Vanishing-Preference Bellman Consistency]
\label{thm:paiqn_consistency}
Suppose target-action selection follows
Eq.~\eqref{eq:preference_target_action}. Then, for all sufficiently large
$t$ such that $\Delta_t\leq1$,
\begin{equation}
\begin{split}
\left\|Q_{t+1}-Q^\star\right\|_\infty
\leq{}&
\gamma\left\|Q_t-Q^\star\right\|_\infty\\
&+
\frac{\gamma D}{K}
+
2\gamma D\Delta_t.
\end{split}
\label{eq:paiqn_consistency}
\end{equation}
Hence, the additional Bellman error induced by preference weighting is
bounded by $2\gamma D\Delta_t$ and vanishes as
$\Delta_t\rightarrow0$.
\end{theorem}

\begin{proof}
For $\Delta_t\leq1$, clipping in Eq.~\eqref{eq:training_weights} is inactive.
Since $\sum_{k=1}^{K}x_{t,k}=1$, normalization gives
\[
w_{t,k}
=
\frac{1-\Delta_t}{K}
+
\Delta_t x_{t,k}.
\]

Define the uniform finite-grid estimate
\[
Q_{t,K}(s,a)
=
\frac{1}{K}
\sum_{k=1}^{K}
Z_{t,\tau_k}(s,a),
\]
and the corresponding preference-weighted score
\[
Q_{t,w}(s,a)
=
\sum_{k=1}^{K}
w_{t,k}Z_{t,\tau_k}(s,a).
\]
Substituting the expression of $w_{t,k}$ yields
\[
Q_{t,w}(s,a)
=
(1-\Delta_t)Q_{t,K}(s,a)
+
\Delta_t
\sum_{k=1}^{K}
x_{t,k}Z_{t,\tau_k}(s,a),
\]
\[
\begin{aligned}
&Q_{t,w}(s,a)-Q_{t,K}(s,a)\\
&\quad=
\Delta_t
\Bigg[
\sum_{k=1}^{K}
x_{t,k}Z_{t,\tau_k}(s,a)
-
Q_{t,K}(s,a)
\Bigg].
\end{aligned}
\]
Both terms inside the brackets are convex combinations of the same quantile
values. Since the discounted return distribution lies within an interval of
width at most $D$, their difference is bounded by $D$, giving
\begin{equation}
\left|
Q_{t,w}(s,a)-Q_{t,K}(s,a)
\right|
\leq
D\Delta_t.
\label{eq:preference_error}
\end{equation}

Next, recall that
$Q_t(s,a)=\int_0^1 Z_{t,\tau}(s,a)\,\mathrm{d}\tau$.
For the midpoint grid
$\tau_k=(k-\frac{1}{2})/K$, monotonicity of the quantile function implies
that, on each interval $[(k-1)/K,k/K]$,
\[
\begin{aligned}
&
\left|
\frac{1}{K}Z_{t,\tau_k}(s,a)
-
\int_{I_k}Z_{t,\tau}(s,a)\,\mathrm{d}\tau
\right|\\
&\quad\leq
\frac{1}{2K}
\Big[
Z_{t,k/K}(s,a)
-
Z_{t,(k-1)/K}(s,a)
\Big].
\end{aligned}
\]
Summing over all intervals and using the bounded return range yields
\begin{equation}
\begin{split}
\left|
Q_{t,K}(s,a)-Q_t(s,a)
\right|
&\leq
\frac{
Z_{t,1}(s,a)-Z_{t,0}(s,a)
}{2K}\\
&\leq
\frac{D}{2K}.
\end{split}
\label{eq:quantile_error}
\end{equation}
Combining Eqs.~\eqref{eq:preference_error} and
\eqref{eq:quantile_error} gives, uniformly over actions,
\begin{equation}
\left|
Q_{t,w}(s,a)-Q_t(s,a)
\right|
\leq
\frac{D}{2K}+D\Delta_t.
\label{eq:combined_error}
\end{equation}

Let
\(
a_t^\star(s')
=
\operatorname*{argmax}_{a\in\mathcal A}
Q_t(s',a)
\)
denote the mean-greedy action, while
$\hat a_t^{\mathrm{tar}}(s')$ is selected by the preference-weighted score in
Eq.~\eqref{eq:preference_target_action}.
Since $\hat a_t^{\mathrm{tar}}(s')$ maximizes $Q_{t,w}(s',a)$,
Eq.~\eqref{eq:combined_error} implies
\begin{equation}
\begin{split}
&Q_t(s',a_t^\star(s'))
-
Q_t(s',\hat a_t^{\mathrm{tar}}(s'))\\
&\qquad\leq
2\left(
\frac{D}{2K}+D\Delta_t
\right)
=
\frac{D}{K}+2D\Delta_t.
\end{split}
\label{eq:action_error}
\end{equation}
Thus, the finite quantile grid contributes at most $D/K$ to the
target-action error, whereas preference weighting contributes at most
$2D\Delta_t$.

Taking expectations in the Bellman update, the latter action-selection error
is discounted by $\gamma$. Applying the $\gamma$-contraction of the standard
Bellman optimality operator therefore gives
\[
\left\|Q_{t+1}-Q^\star\right\|_\infty
\leq
\gamma\left\|Q_t-Q^\star\right\|_\infty
+
\frac{\gamma D}{K}
+
2\gamma D\Delta_t,
\]
which proves Eq.~\eqref{eq:paiqn_consistency}.
Finally, Eq.~\eqref{eq:training_weights} gives
$\Delta_t=\Delta_0\max(t,1)^{-(1+\eta)}$ with $\eta>0$, so
$\Delta_t\rightarrow0$. Consequently, the additional Bellman error
$2\gamma D\Delta_t$ induced by preference weighting vanishes asymptotically.
\end{proof}

Theorem~\ref{thm:paiqn_consistency} shows that preference-guided exploration
can diversify action evaluation during training without introducing a
persistent bias into the underlying risk-neutral Bellman objective. As
$\Delta_t\rightarrow0$, the preference-induced error vanishes, so the
asymptotic behavior is governed only by the approximation of the return mean.
In particular, when this approximation error vanishes,
$Q_t\rightarrow Q^\star$. For a fixed grid of $K$ midpoint quantiles, the
remaining finite-grid error satisfies
\begin{equation}
\limsup_{t\rightarrow\infty}
\left\|Q_t-Q^\star\right\|_\infty
\leq
\frac{\gamma D}{K(1-\gamma)}.
\label{eq:paiqn_neighborhood}
\end{equation}
Thus, the proposed preference mechanism provides diverse training-time
action preferences while preserving asymptotic consistency with the
risk-neutral Bellman solution up to the finite quantile-grid approximation.

\subsection{Deployment-Time Robust Preference Evaluation}\label{subsec:deployment}

Training-time preference sampling encourages broad behavioral coverage,
whereas deployment requires a deterministic action rule. Given a state $s$,
we therefore disable both random-action exploration and random preference
sampling, and query the trained IQN once at $K_{\mathrm{inf}}=64$ fixed
midpoint quantiles for all candidate actions.

At test time, we allow the quantile weights to vary within a KL neighborhood
of the uniform weights. For each action, its robust value is
\begin{equation}
\begin{aligned}
Q_\rho(s,a)=
\min_{w_1,\ldots,w_{K_{\mathrm{inf}}}}\quad
&\sum_{k=1}^{K_{\mathrm{inf}}}w_kZ_{\tau_k}(s,a)\\
\text{s.t.}\quad
&w_k\geq0,\qquad \sum_{k=1}^{K_{\mathrm{inf}}}w_k=1,\\
&\sum_{k=1}^{K_{\mathrm{inf}}}w_k
\log(K_{\mathrm{inf}}w_k)\leq\rho,
\end{aligned}
\label{eq:robust_value}
\end{equation}
where $\tau_k=(k-\tfrac{1}{2})/K_{\mathrm{inf}}$ and $\rho\geq0$ controls the
allowed deviation from uniform weighting. Thus, $\rho=0$ recovers the midpoint
approximation of the expected return, whereas a larger $\rho$ allows more mass
to be assigned to low predicted returns. The same radius is used for all
candidate actions.

The inner minimization in Eq.~\eqref{eq:robust_value} is convex and is solved
independently for each action. For nonconstant quantile predictions and an
active KL constraint, its globally optimal weights follow from the KKT
conditions~\cite{vandenberghe2004convex} for KL-divergence-constrained robust optimization~\cite{ben2013robust}. Using the nonnegative inverse temperature $\lambda_{s,a}$, they
can be written as follows, where
$z_{a,k}$ denotes $Z_{\tau_k}(s,a)$ for short:
\begin{equation}
\begin{gathered}
w_k^*(s,a)
=
\frac{\exp(-\lambda_{s,a}z_{a,k})}
{\displaystyle\sum_{j=1}^{K_{\mathrm{inf}}}
\exp(-\lambda_{s,a}z_{a,j})},
\\[3pt]
\sum_{k=1}^{K_{\mathrm{inf}}}
w_k^*(s,a)
\log\!\left(
K_{\mathrm{inf}}w_k^*(s,a)
\right)
=
\rho .
\end{gathered}
\label{eq:robust_weights}
\end{equation}
The second line is a scalar root equation. Its left-hand side increases
monotonically from zero as $\lambda_{s,a}$ increases, so we first bracket the
root by doubling an upper endpoint and then apply bisection. For numerical
stability, the quantile predictions are shifted and scaled to $[0,1]$ before
the search; this only reparameterizes $\lambda_{s,a}$ and leaves the minimizing
weights unchanged. 

At $\rho=0$, or when all quantile predictions for an action are equal, the
solver returns uniform weights. If the radius is large enough to permit a
distribution supported only on the minimum-valued prediction or predictions,
the robust score has saturated at that minimum and the solver assigns equal
mass to those entries. These cases avoid an unnecessary or ill-conditioned
root search.

After solving Eq.~\eqref{eq:robust_weights} for every action, we substitute
each action-specific weight vector directly into Eq.~\eqref{eq:robust_value}
and select
\begin{equation}
\pi_{\mathrm{PAIQN}}(s)
=\operatorname*{argmax}_{a\in\mathcal A}Q_\rho(s,a).
\label{eq:paiqn_policy}
\end{equation}
Beyond the IQN forward evaluation, the robust reweighting introduces a
computational complexity of
$O(|\mathcal{A}|K_{\mathrm{inf}}L)$, where $L$ denotes the number of
bisection iterations used to solve for the action-specific inverse
temperature.

\section{Experimental Results}

\begingroup

\newcommand{\mixmstd}[2]{%
    \ensuremath{#1\,{\scriptstyle\pm\,#2}}}
\newcommand{\bestmstd}[2]{%
    \ensuremath{\mathbf{#1}\,{\scriptstyle\boldsymbol{\pm}\,\mathbf{#2}}}}
\newcommand{\secondmstd}[2]{%
    \ensuremath{\underline{#1\,{\scriptstyle\pm\,#2}}}}

\begin{table*}[ht]
\centering
\vspace{0.2cm}
\caption{\textbf{Marine ASV Navigation Task:}
Quantitative comparison across different
environment difficulty levels. The best-performing results are highlighted
in \textbf{bold} and the second best results are highlighted with an \underline{underline}.}
\label{tab:asv_main}

\footnotesize
\setlength{\tabcolsep}{1.6pt}
\renewcommand{\arraystretch}{0.90}

\begin{tabular*}{\textwidth}
{@{\extracolsep{\fill}}lllrrrrrr@{}}
\toprule
\textbf{Difficulty}
& \textbf{Training Method}
& \textbf{Preference during deployment}
& \shortstack{\textbf{SR} (\%) $\uparrow$}
& \shortstack{\textbf{CR} (\%) $\downarrow$}
& \shortstack{\textbf{TR} (\%) $\downarrow$}
& \shortstack{\textbf{Reward} $\uparrow$}
& \shortstack{\textbf{Succ. time} (s) $\downarrow$}
& \shortstack{\textbf{Succ. energy} $\downarrow$} \\
\midrule

\multirow{14}{*}{Easy}
& DQN & Greedy
& \mixmstd{87.3}{4.6}
& \mixmstd{12.7}{4.6}
& \bestmstd{0.0}{0.0}
& \mixmstd{66.2}{6.5}
& \mixmstd{35.3}{0.4}
& \mixmstd{104.2}{1.1} \\

\cmidrule(lr){2-9}

& \multirow{6}{*}{IQN}
& CVaR ($\alpha=.25$)
& \mixmstd{30.7}{5.8}
& \mixmstd{68.0}{5.3}
& \secondmstd{1.3}{1.2}
& \mixmstd{-117.2}{20.7}
& \mixmstd{83.1}{25.0}
& \secondmstd{76.4}{21.2} \\

& & CVaR ($\alpha=.50$)
& \mixmstd{54.7}{5.0}
& \mixmstd{45.3}{5.0}
& \bestmstd{0.0}{0.0}
& \mixmstd{-39.3}{12.5}
& \mixmstd{68.6}{4.2}
& \mixmstd{82.0}{1.9} \\

& & CVaR ($\alpha=.75$)
& \mixmstd{92.0}{2.0}
& \mixmstd{8.0}{2.0}
& \bestmstd{0.0}{0.0}
& \mixmstd{75.3}{3.7}
& \mixmstd{35.4}{0.6}
& \mixmstd{93.8}{1.4} \\

& & Risk-neutral ($\alpha=1$)
& \mixmstd{92.7}{4.2}
& \mixmstd{7.3}{4.2}
& \bestmstd{0.0}{0.0}
& \mixmstd{75.8}{6.0}
& \mixmstd{34.9}{0.1}
& \mixmstd{98.7}{0.4} \\

& & Distance-adaptive
& \mixmstd{92.7}{2.3}
& \mixmstd{7.3}{2.3}
& \bestmstd{0.0}{0.0}
& \mixmstd{76.1}{3.5}
& \mixmstd{35.1}{0.1}
& \mixmstd{98.9}{0.6} \\

& & RTV
& \mixmstd{89.3}{4.2}
& \mixmstd{10.7}{4.2}
& \bestmstd{0.0}{0.0}
& \mixmstd{68.5}{6.6}
& \mixmstd{35.8}{0.8}
& \mixmstd{100.5}{2.6} \\

\cmidrule(lr){2-9}

& \multirow{7}{*}{PAIQN (ours)}
& CVaR ($\alpha=.25$)
& \mixmstd{8.7}{3.1}
& \mixmstd{88.0}{4.0}
& \mixmstd{3.3}{4.2}
& \mixmstd{-190.6}{28.7}
& \mixmstd{145.7}{26.0}
& \bestmstd{57.5}{0.7} \\

& & CVaR ($\alpha=.50$)
& \mixmstd{94.7}{7.6}
& \mixmstd{5.3}{7.6}
& \bestmstd{0.0}{0.0}
& \mixmstd{76.6}{10.7}
& \mixmstd{35.0}{1.0}
& \mixmstd{91.5}{2.9} \\

& & CVaR ($\alpha=.75$)
& \secondmstd{96.7}{3.1}
& \secondmstd{4.0}{4.0}
& \bestmstd{0.0}{0.0}
& \bestmstd{81.4}{5.0}
& \secondmstd{34.2}{0.4}
& \mixmstd{91.7}{1.1} \\

& & Risk-neutral ($\alpha=1$)
& \mixmstd{96.0}{3.5}
& \secondmstd{4.0}{3.5}
& \bestmstd{0.0}{0.0}
& \mixmstd{80.8}{5.2}
& \bestmstd{34.0}{0.5}
& \mixmstd{93.8}{1.4} \\

& & Distance-adaptive
& \secondmstd{96.7}{1.2}
& \secondmstd{4.0}{2.0}
& \bestmstd{0.0}{0.0}
& \secondmstd{81.1}{2.5}
& \mixmstd{34.5}{0.3}
& \mixmstd{95.3}{1.1} \\

& & RTV
& \mixmstd{96.0}{2.0}
& \mixmstd{4.7}{3.1}
& \bestmstd{0.0}{0.0}
& \mixmstd{76.4}{3.1}
& \mixmstd{36.0}{0.6}
& \mixmstd{98.2}{0.7} \\

& & Robust Preference (ours)
& \bestmstd{97.3}{3.1}
& \bestmstd{2.7}{3.1}
& \bestmstd{0.0}{0.0}
& \mixmstd{80.9}{5.3}
& \mixmstd{34.9}{0.3}
& \mixmstd{92.2}{0.5} \\

\midrule

\multirow{14}{*}{Medium}
& DQN & Greedy
& \mixmstd{78.7}{7.6}
& \mixmstd{22.0}{7.2}
& \bestmstd{0.0}{0.0}
& \mixmstd{54.1}{11.5}
& \bestmstd{35.0}{0.5}
& \mixmstd{105.8}{2.2} \\

\cmidrule(lr){2-9}

& \multirow{6}{*}{IQN}
& CVaR ($\alpha=.25$)
& \mixmstd{23.3}{2.3}
& \mixmstd{76.7}{2.3}
& \bestmstd{0.0}{0.0}
& \mixmstd{-84.0}{14.6}
& \mixmstd{84.4}{21.9}
& \secondmstd{72.3}{3.2} \\

& & CVaR ($\alpha=.50$)
& \mixmstd{52.0}{4.0}
& \mixmstd{48.0}{4.0}
& \bestmstd{0.0}{0.0}
& \mixmstd{-7.9}{4.4}
& \mixmstd{54.5}{3.7}
& \mixmstd{88.2}{5.5} \\

& & CVaR ($\alpha=.75$)
& \mixmstd{80.0}{6.0}
& \mixmstd{20.0}{6.0}
& \bestmstd{0.0}{0.0}
& \mixmstd{57.8}{6.0}
& \mixmstd{36.1}{0.6}
& \mixmstd{98.8}{1.9} \\

& & Risk-neutral ($\alpha=1$)
& \mixmstd{80.7}{10.3}
& \mixmstd{19.3}{10.3}
& \bestmstd{0.0}{0.0}
& \mixmstd{59.2}{11.3}
& \mixmstd{35.5}{0.5}
& \mixmstd{104.0}{1.4} \\

& & Distance-adaptive
& \mixmstd{80.7}{9.9}
& \mixmstd{20.0}{10.4}
& \bestmstd{0.0}{0.0}
& \mixmstd{57.6}{11.6}
& \mixmstd{36.2}{1.1}
& \mixmstd{105.9}{2.6} \\

& & RTV
& \mixmstd{77.3}{4.2}
& \mixmstd{22.7}{4.2}
& \bestmstd{0.0}{0.0}
& \mixmstd{46.3}{3.0}
& \mixmstd{39.4}{2.3}
& \mixmstd{114.7}{7.2} \\

\cmidrule(lr){2-9}

& \multirow{7}{*}{PAIQN (ours)}
& CVaR ($\alpha=.25$)
& \mixmstd{10.7}{4.6}
& \mixmstd{89.3}{4.6}
& \bestmstd{0.0}{0.0}
& \mixmstd{-109.5}{5.7}
& \mixmstd{98.1}{20.8}
& \bestmstd{57.3}{1.0} \\

& & CVaR ($\alpha=.50$)
& \mixmstd{88.0}{6.0}
& \mixmstd{11.3}{7.0}
& \secondmstd{0.7}{1.2}
& \mixmstd{59.3}{5.8}
& \mixmstd{36.4}{0.6}
& \mixmstd{97.7}{2.0} \\

& & CVaR ($\alpha=.75$)
& \mixmstd{88.7}{8.3}
& \mixmstd{11.3}{8.3}
& \bestmstd{0.0}{0.0}
& \secondmstd{68.7}{12.5}
& \secondmstd{35.1}{1.4}
& \mixmstd{96.8}{4.1} \\

& & Risk-neutral ($\alpha=1$)
& \mixmstd{86.7}{6.1}
& \mixmstd{13.3}{6.1}
& \bestmstd{0.0}{0.0}
& \mixmstd{65.4}{9.4}
& \bestmstd{35.0}{0.7}
& \mixmstd{97.8}{1.2} \\

& & Distance-adaptive
& \secondmstd{89.3}{4.2}
& \secondmstd{10.7}{4.2}
& \bestmstd{0.0}{0.0}
& \mixmstd{67.6}{6.0}
& \mixmstd{35.9}{1.4}
& \mixmstd{101.2}{3.5} \\

& & RTV
& \mixmstd{85.3}{5.8}
& \mixmstd{14.7}{5.8}
& \bestmstd{0.0}{0.0}
& \mixmstd{53.3}{3.4}
& \mixmstd{39.6}{1.5}
& \mixmstd{112.5}{4.2} \\

& & Robust Preference (ours)
& \bestmstd{90.0}{6.9}
& \bestmstd{10.0}{6.9}
& \bestmstd{0.0}{0.0}
& \bestmstd{69.2}{9.0}
& \mixmstd{35.6}{1.2}
& \mixmstd{96.4}{2.8} \\

\midrule

\multirow{14}{*}{Hard}
& DQN & Greedy
& \mixmstd{66.7}{3.1}
& \mixmstd{33.3}{3.1}
& \bestmstd{0.0}{0.0}
& \mixmstd{35.2}{5.2}
& \bestmstd{34.5}{0.6}
& \mixmstd{106.1}{2.4} \\

\cmidrule(lr){2-9}

& \multirow{6}{*}{IQN}
& CVaR ($\alpha=.25$)
& \mixmstd{17.3}{3.1}
& \mixmstd{81.3}{2.3}
& \mixmstd{1.3}{1.2}
& \mixmstd{-92.7}{11.7}
& \mixmstd{63.1}{4.9}
& \secondmstd{78.2}{13.1} \\

& & CVaR ($\alpha=.50$)
& \mixmstd{43.3}{5.0}
& \mixmstd{56.7}{5.0}
& \bestmstd{0.0}{0.0}
& \mixmstd{-7.1}{7.9}
& \mixmstd{46.6}{0.6}
& \mixmstd{93.8}{6.1} \\

& & CVaR ($\alpha=.75$)
& \mixmstd{67.3}{8.1}
& \mixmstd{33.3}{8.3}
& \bestmstd{0.0}{0.0}
& \mixmstd{39.8}{12.0}
& \mixmstd{38.3}{2.5}
& \mixmstd{105.8}{6.6} \\

& & Risk-neutral ($\alpha=1$)
& \mixmstd{73.3}{9.0}
& \mixmstd{27.3}{9.0}
& \bestmstd{0.0}{0.0}
& \mixmstd{48.0}{13.6}
& \mixmstd{36.3}{0.5}
& \mixmstd{107.6}{1.7} \\

& & Distance-adaptive
& \mixmstd{74.0}{8.0}
& \mixmstd{26.7}{8.1}
& \bestmstd{0.0}{0.0}
& \mixmstd{45.9}{12.2}
& \mixmstd{37.7}{0.7}
& \mixmstd{111.6}{2.8} \\

& & RTV
& \mixmstd{65.3}{4.2}
& \mixmstd{32.0}{3.5}
& \mixmstd{3.3}{1.2}
& \mixmstd{-13.4}{24.0}
& \mixmstd{40.2}{2.8}
& \mixmstd{119.2}{7.6} \\

\cmidrule(lr){2-9}

& \multirow{7}{*}{PAIQN (ours)}
& CVaR ($\alpha=.25$)
& \mixmstd{8.7}{3.1}
& \mixmstd{90.7}{3.1}
& \secondmstd{0.7}{1.2}
& \mixmstd{-114.3}{10.1}
& \mixmstd{80.4}{16.2}
& \bestmstd{77.4}{17.4} \\

& & CVaR ($\alpha=.50$)
& \secondmstd{82.7}{5.0}
& \secondmstd{17.3}{5.0}
& \bestmstd{0.0}{0.0}
& \mixmstd{53.5}{4.1}
& \mixmstd{38.7}{1.8}
& \mixmstd{104.7}{4.9} \\

& & CVaR ($\alpha=.75$)
& \mixmstd{81.3}{6.4}
& \mixmstd{18.0}{7.2}
& \secondmstd{0.7}{1.2}
& \mixmstd{48.0}{4.9}
& \secondmstd{35.8}{1.2}
& \mixmstd{99.2}{3.4} \\

& & Risk-neutral ($\alpha=1$)
& \secondmstd{82.7}{4.2}
& \mixmstd{18.0}{4.0}
& \bestmstd{0.0}{0.0}
& \secondmstd{55.6}{7.3}
& \mixmstd{37.3}{0.7}
& \mixmstd{106.2}{2.3} \\

& & Distance-adaptive
& \mixmstd{80.0}{6.9}
& \mixmstd{20.0}{6.9}
& \bestmstd{0.0}{0.0}
& \mixmstd{51.7}{10.2}
& \mixmstd{36.9}{0.2}
& \mixmstd{104.8}{0.2} \\

& & RTV
& \mixmstd{72.0}{10.0}
& \mixmstd{26.7}{10.1}
& \mixmstd{1.3}{1.2}
& \mixmstd{22.6}{16.0}
& \mixmstd{38.8}{0.4}
& \mixmstd{110.4}{1.6} \\

& & Robust Preference (ours)
& \bestmstd{86.7}{2.3}
& \bestmstd{13.3}{2.3}
& \bestmstd{0.0}{0.0}
& \bestmstd{58.6}{4.9}
& \mixmstd{38.9}{2.8}
& \mixmstd{107.2}{8.3} \\

\bottomrule
\end{tabular*}
\vspace{-0.4cm}
\end{table*}
\endgroup

\subsection{Implementation Details}

\subsubsection{Environments Simulation}

To evaluate both the navigation performance of PAIQN and its generalization
across different sources of risk and uncertainty, we consider two
sensor-based environments adapted from prior work: a marine ASV
task~\cite{lin2023robust} with obstacles and unobservable currents, and a
cluttered indoor drone task~\cite{liu2022adaptive} characterized by
occlusions and limited maneuvering space. Both environments involve map-free
navigation under randomized layouts and environmental disturbances, while
differing in robotic platforms, sensing modalities, and dominant uncertainty
sources, thereby providing complementary testbeds for evaluating robustness
and generalization under partial observability.

In the marine environment, the vehicle perceives surrounding obstacles using an $11$-beam forward-looking sonar covering a $2\pi/3$ rad sector with an effective range of approximately $10\,\mathrm{m}$. Motion and goal-related information is obtained from DVL, IMU, compass, and GPS measurements. Environmental vortices are introduced to perturb vehicle motion, while the obstacle and disturbance layouts are re-initialized across episodes.
For the indoor drone environment, four single-beam range sensors oriented along the body-frame $\pm x$ and $\pm y$ directions provide sparse obstacle measurements with a maximum range of about $4\,\mathrm{m}$. The obstacle layout is likewise randomized between episodes.

\subsubsection{Hyperparameter Settings}
For reproducibility and numerical stability, we follow the training configuration adopted in previous work~\cite{lin2023robust,liu2022adaptive}. 
For ASV navigation policy learning in marine environment, the exploration rate is linearly annealed from \(\epsilon_0=1\) to \(\epsilon_f=0.05\) over the first \(f_\epsilon=0.1\) fraction of training and remains at \(0.05\) thereafter. 
A grid search indicated that performance is insensitive to the initial value \(\Delta_0\); we therefore fix it to a sufficiently large value, \(\Delta_0=5\times10^3\) with decay rate \(\eta=0.001\) in Eq.~\eqref{eq:training_weights}. The Dirichlet concentration $b$ is set to \(0.05\) to induce more diverse and non-uniform preferences in the quantile weights. After training, the radius $\rho=5\times10^{-3}$ in Eq.\eqref{eq:robust_value} is selected using separate validation scenarios and then fixed throughout test-time deployment. We use a stop tolerance $10^{-10}$ and at most $L=160$ bisection iterations.
For indoor drone navigation, the radius decay of preference exploration is proportionally rescaled to the shorter training horizon, while retaining \(\Delta_0=5\times10^3\), \(\eta=0.001\), and \(b=0.05\).
For Robust Preference during testing, \(\rho=1\times10^{-4}\) is selected on separate validation scenarios, with tolerance \(10^{-9}\) and at most 100 bisection iterations.

\subsection{Baseline and Metrics}
We compare PAIQN against a standard RL baseline and representative
DistRL-based methods that adjust risk preferences during action evaluation.
Specifically, the baselines include: (1) DQN~\cite{Mnih2015HumanlevelCT} as
the standard value-based RL method; (2) risk-neutral
IQN~\cite{dabney2018implicit} with $\mathrm{CVaR}=1.0$;
(3) fixed-risk IQN with
$\mathrm{CVaR}\in\{0.75,0.50,0.25\}$; and two adaptive
risk-preference adjustment methods based on DistRL:
(4) RTV~\cite{liu2022adaptive}, which adapts the CVaR level using the
lower-tail conditional variance of the predicted return distribution with
exponentially weighted forecasting; and
(5) Perceived Distance~\cite{lin2023robust}, which adjusts the CVaR level
according to the distance to the nearest obstacle.
To comprehensively evaluate PAIQN against existing risk-sensitive strategies,
we report six metrics covering task performance, safety, and execution
efficiency: success rate (SR), collision rate (CR), timeout rate (TR), final reward, average
completion time (Succ. time), and average energy consumption (Succ. energy). For consistent evaluation, out-of-boundary cases are counted as collisions. Completion time and energy
consumption are averaged over successful episodes only. These two metrics
characterize execution efficiency and are also affected by the length and
smoothness of the resulting trajectories.

\subsection{Results and Discussion}

\begin{figure}[t]
\centering  
\vspace{0.2cm}
\includegraphics[width=0.98\linewidth]{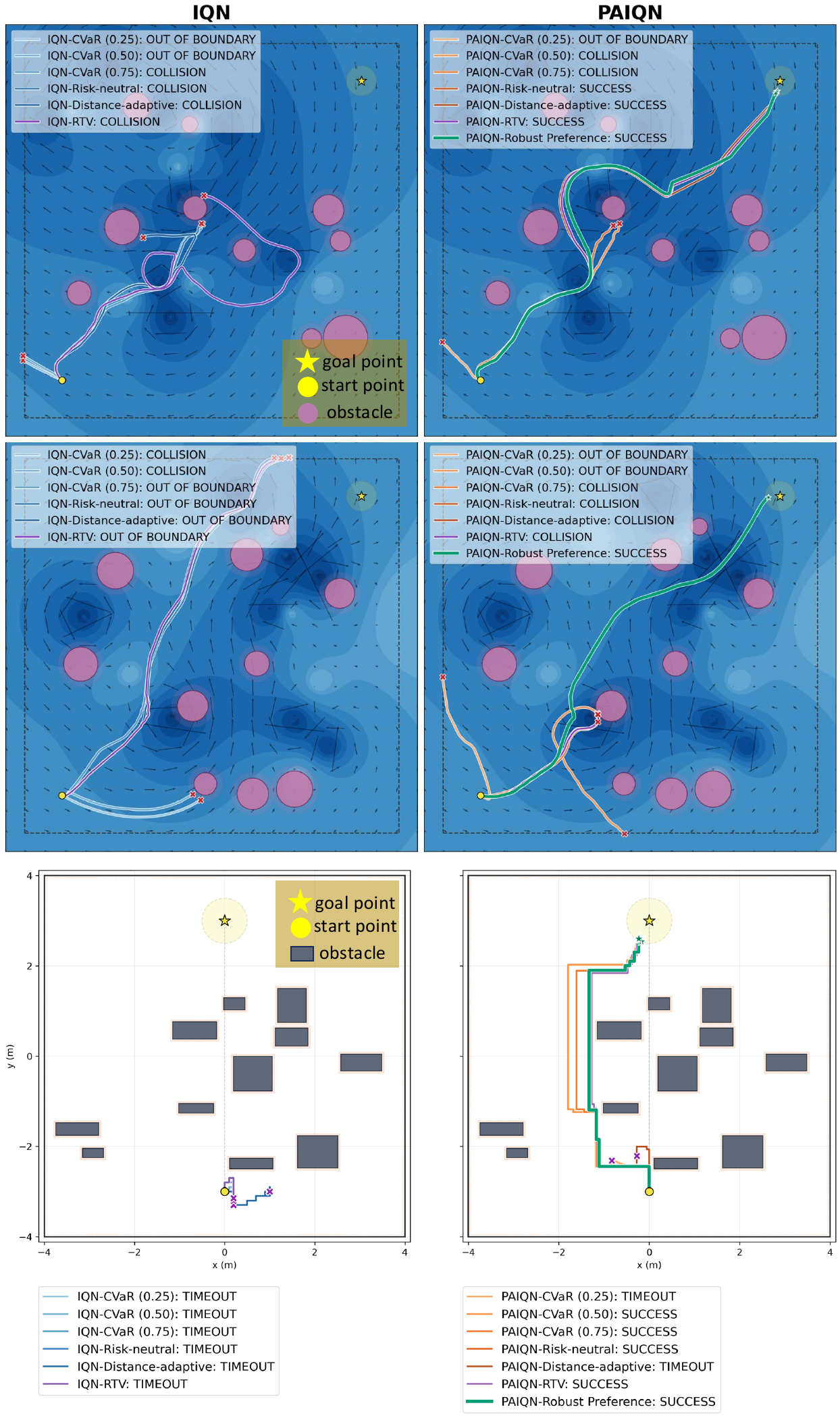}
\vspace{-0.25cm}
\caption{\textbf{Qualitative trajectory results} (first two rows: marine navigation; last row:
indoor navigation). Aligning with Table~\ref{tab:asv_main} and~\ref{tab:drone_results}, PAIQN demonstrates superior performance across different risk preferences, while Robust Preference further improves performance in complex scenarios.}
\vspace{-0.6cm}
\label{fig3}
\end{figure}

\subsubsection{Environmental Settings and Risk Preference}
We conduct evaluation using the test settings supported by the original simulation environments. For the marine task, we consider three progressively more challenging settings: \textit{Easy} (4 vortices and 6 obstacles), \textit{Medium} (6 vortices and 8 obstacles), and \textit{Hard} (8 vortices and 10 obstacles). For each difficulty level, we independently generate three sets of 50 distinct environment layouts to reduce the effect of randomness. For the indoor task, we randomize the environment layout with up to 11 obstacles and evaluate each method on three independently generated sets of 100 environments.
Consistent with the default risk configuration during training, the marine task uses a fixed risk-neutral policy for action selection, whereas the indoor task samples a CVaR level at the beginning of each episode. 
In contrast, PAIQN employs progressively annealed preference-guided exploration, exposing the agent to diverse preferences at each decision while gradually reducing exploration to limit the resulting learning bias.

\begingroup

\newcommand{\mixmstd}[2]{%
    \ensuremath{#1\,{\scriptstyle\pm\,#2}}}
\newcommand{\bestmstd}[2]{%
    \ensuremath{\mathbf{#1}\,{\scriptstyle\pm\,\mathbf{#2}}}}
\newcommand{\secondmstd}[2]{%
    \ensuremath{\underline{#1\,{\scriptstyle\pm\,#2}}}}

\begin{table}[ht]
\centering
\caption{\textbf{Indoor Drone Navigation Task}: quantitative simulation performance under different risk preference.}
\label{tab:drone_results}

\footnotesize
\setlength{\tabcolsep}{2.5pt}
\renewcommand{\arraystretch}{1.04}

\begin{tabular}{@{}llcc@{}}
\toprule
\textbf{Method}
& \textbf{Preference}
& \textbf{SR} (\%) $\uparrow$
& \textbf{CR} (\%) $\downarrow$ \\
\midrule

DQN
& Greedy
& \mixmstd{40.17}{2.31}
& \mixmstd{23.67}{0.76} \\

\midrule

\multirow{3}{*}{IQN}
& Best Fixed CVaR ($\alpha=.50$)
& \mixmstd{70.17}{2.36}
& \mixmstd{6.17}{1.53} \\

& RTV
& \mixmstd{68.67}{1.61}
& \secondmstd{6.00}{1.00} \\

& Distance-adaptive
& \mixmstd{63.33}{3.33}
& \mixmstd{8.00}{1.32} \\

\midrule

\multirow{4}{*}{PAIQN (ours)}
& Best Fixed CVaR ($\alpha=1$)
& \secondmstd{84.00}{3.04}
& \mixmstd{7.50}{0.50} \\

& RTV
& \secondmstd{84.00}{3.04}
& \mixmstd{7.33}{0.76} \\

& Distance-adaptive
& \mixmstd{75.67}{5.48}
& \bestmstd{5.17}{1.89} \\

& Robust Preference (ours)
& \bestmstd{84.33}{3.18}
& \mixmstd{7.00}{0.87} \\

\bottomrule
\end{tabular}
\vspace{-0.2cm}
\end{table}
\endgroup

\subsubsection{Training Preference Benefits: PAIQN \textsc{vs.} Baseline Approaches}
The qualitative trajectories of RL agents are visualized in Fig.~\ref{fig3} in representative complex environments characterized by substantial environmental uncertainty.
As shown in Table~\ref{tab:asv_main}, PAIQN consistently improves both
success and safety across the \textbf{marine environments} under different
risk preferences, with the exception of the highly conservative
$\mathrm{CVaR}=0.25$ setting. The gains are particularly pronounced in the
\textit{Hard} environment with complex vortex dynamics and collision hazards.
This suggests that PAIQN's diverse training preferences are most beneficial
when deployment-time evaluation preserves information across the return
distribution. In contrast, $\mathrm{CVaR}=0.25$ uses only the lowest quarter,
overemphasizing tail risk and discarding the remaining quantiles, thereby
limiting the benefit of preference-diverse training.
As illustrated in Table~\ref{tab:drone_results}, in the \textbf{indoor environment}, PAIQN substantially
outperforms IQN trained with episode-level CVaR sampling. The comparable
performance of several deployment-time strategies may result from the sensing
and environmental configuration, while the overall gains further support the
generalization of PAIQN across environments and robotic platforms.

\subsubsection{Preference Benefits at Deployment: PAIQN with Robust Preference}

As reported in Table~\ref{tab:asv_main}, applying Robust Preference to PAIQN at deployment further improves success rates and reduces collisions across all three difficulty levels in the marine environment.
The gains are particularly pronounced in
the Hard setting, highlighting the benefit of DRO-based robust preference
evaluation during deployment. These results also reveal the limitation of
relying on a single fixed risk preference: under complex uncertainty, the
appropriate degree of conservativeness may vary substantially across states.
Moreover, adapting risk solely from partially observable sensory signals can
be less effective when the observations provide incomplete evidence of latent
hazards. Robustifying the risk preference itself therefore provides additional
resilience beyond fixed or signal-driven risk adaptation.
Importantly, the additional computational cost of solving Eq.~\eqref{eq:robust_value} is about $12\,\mathrm{ms}$, preserving the efficiency required for real-time decision-making.


\section{Conclusion}

We presented PAIQN, a preference-augmented distributional reinforcement learning
framework for risk-sensitive autonomous navigation under partial observability
and latent environmental uncertainty. Rather than restricting risk preference
to a scalar CVaR level, PAIQN represents it through quantile-weight
distributions. During training, progressively annealed preference-guided
exploration exposes the agent to diverse action preferences, while our
theoretical analysis shows that the resulting preference-induced Bellman
deviation vanishes as the perturbation is annealed. At deployment, we formulate
risk adaptation as a distributionally robust optimization problem over
quantile weights, enabling conservative action evaluation without
mapping task-specific risk indicators to a CVaR level. Experiments across
marine ASV and indoor drone navigation tasks demonstrate improved safety and
task success under different sources of uncertainty.

Our evaluation is currently limited to simulation, and the deployment-time
ambiguity radius is selected on validation scenarios and fixed thereafter.
Future work will explore online adaptation of the ambiguity set, validation on
physical robots, and broader navigation settings and action spaces.
We expect preference-augmented distributional reasoning to support robust
decision-making when environmental risk cannot be fully inferred from
instantaneous observations.












\bigskip
\bibliography{reference}
\bibliographystyle{IEEEtran}

\end{document}